\documentclass[11pt]{article}

\usepackage[preprint]{acl}

\usepackage{times}
\usepackage{latexsym}
\usepackage{amssymb}
\usepackage{mathtools}
\usepackage{appendix}
\usepackage[table]{xcolor}
\newcommand{\basegray}[1]{\cellcolor{gray!20}#1}

\DeclarePairedDelimiterX{\infdivx}[2]{(}{)}{%
  #1\;\delimsize\|\;#2%
}

\usepackage[T1]{fontenc}

\usepackage[utf8]{inputenc}

\usepackage{microtype}

\usepackage{inconsolata}

\usepackage{graphicx}

\usepackage{amsmath}
\usepackage{amssymb}
\usepackage{booktabs}
\usepackage{multirow}
\usepackage{wrapfig}

\newcommand\methodname{\textcolor{black}{OPD+}}
\newcommand{\opd}{\textsc{OPD}}

\title{\texttt{OPD+}: Rethinking the Advantage Design for On-Policy Distillation}

\author{Hanyang Zhao$^*$$^1$, Haoxian Chen$^*$$^2$, Han Lin$^3$, Genta Indra Winata$^4$, \\
\textbf{David Yao$^1$, Wenpin Tang$^1$} \\
$^*$Equal Contribution\quad$^1$Columbia University$\quad$$^2$Amazon$\quad$$^3$Meta$\quad$$^4$Capital One\\
\texttt{hz2684@columbia.edu} \\}

\begin{document}
\maketitle
\begin{abstract}
On-policy distillation (OPD) is a widely used technique to transfer capabilities from capable teacher language models to the base student models, and can be formulated in a reinforcement learning style objective using student generated rollouts. Yet, despite the divergence reward being dependent on student model likelihood, existing works usually adopt a stop gradient design primarily for stability, which makes the resulting advantage estimation questionable. In this work, we provide a generic optimization framework based on $f$-divergence between the student and teacher, and mathematically revisit whether such design space is valid. We prove that general stop-gradient operation would lead to biased estimates of the reward objective and corresponding gradient for general divergence functions. We propose OPD+, the corrected version of OPD that demonstrates improved performance over the baseline KL approach and also supports the choice of various $f$-divergence. We validate our findings on mathematical reasoning and tool-use benchmarks. 

\end{abstract}

\section{Introduction}


On-policy distillation (\textsc{OPD})~\cite{agarwal2024policy,lu2025onpolicydistillation} is a powerful technique to let student models learn from capable teacher models through dense token-level supervisions. It has also become a crucial part of the post training recipes of modern large language models (LLMs), e.g. in Deepseek V4 \cite{deepseekai2026deepseekv4}, to learn from teacher models without performance degeneracy.

In the design space of OPD\footnote{For OPD implementation, existing approaches either adopt a logit (top-k) distillation or the single token likelihood. We focus on the simpler second approach without the need of additional infrastructure modification as in \citet{deepseekai2026deepseekv4}. }, existing baselines like Tinker \cite{lu2025onpolicydistillation} suggest to adopt the \textit{KL divergence} between student and teacher to achieve the best performance. To derive the loss function, common practice also {\it disables the gradient propagation} of log ratio difference reward between student and teacher primarily for numerical stability, motivated by e.g., target networks in DQN \cite{mnih2013playing}.



However, in this paper, we revisit these two key designs for OPD. Our contributions are mainly two-folded: \textbf{Firstly,} we provide the framework of $f$-divergence on policy distillation and derive the corresponding gradient and loss functions dependent on whether gradients on the reward function is enabled. Our framework enriches the design space of on policy distillation with various $f$-divergence; \textbf{Secondly,} we demonstrate that stopping gradient in reward not only yields mathematically incorrectly gradient estimates but also harms the performance for various $f$-divergence functions. The corrected algorithm, which we name as OPD+, yields strong empirical gains over the baseline in math reasoning and tool use scenarios, and challenges the common belief of the advantage of KL divergence over other statistical distances in on policy distillation.


\section{Preliminaries}
\label{Sec: Preliminaries}
We first revisit the preliminaries of LLMs and \textsc{OPD} as a Reinforcement Learning (RL) objective. More discussions on some other relevant works are provided in Appendix \ref{App: related works}.

\noindent \paragraph{LLMs.} Denote by $y\sim p_{\theta}(\cdot \mid x)$ a full sentence generated by language model $p_{\theta}$ given the context/prompt $x$. For auto-regressive LLMs, we use $p_{\theta}$ for a sequential model parameterized by $\theta$, whose likelihood can be factorized as the product of the next-token distribution over the vocabulary, conditioning on the prompt $x$ and already generated tokens $y_{<n}$, i.e.,
$p_{\theta}(y \mid x)=\prod_{n=1}^{L_y} p_{\theta}(y_n \mid y_{<n}, x)$.

\paragraph{OPD.} For OPD, we set the behavior policy the same as the student policy/model, and we are interested in optimizing a per-token level difference between student $p_{\theta}$ and teacher policy $q$:
$$
\mathbb{E}_{x\sim X,\ y\sim p_{\theta}(\cdot \mid x)} \sum_{n=1}^{L_y}D(p_{\theta}(\cdot \mid y_{<n}, x),q(\cdot \mid y_{<n}, x)),
$$
where $D(p,q)$ denotes a chosen statistical divergence between two probability distributions, and $L_y$ is the length of the generation $y$. 
When choosing $D$ as the KL divergence, i.e., $D(P,Q):=D_{\mathrm{KL}}(P\|Q)
= \sum_{x} p(x)(\log p(x)-\log q(x))$, we have an equivalent loss objective which can be simply estimated by student sample rollouts:
$$
\mathbb{E}_{x\sim X,\ y\sim p_{\theta}(\cdot \mid x)} \sum_{n=1}^{L_y}\log \frac{p_{\theta}(y_n \mid y_{<n}, x)}{q(y_n \mid y_{<n}, x)}.
$$
In practice, the log ratio difference is often `stop grad'ed as a {\em fixed} reward. 
Using a zero discount factor for the advantage function yields the PPO-like loss
$\mathcal{J}_{\text{opd}}(\theta):=$
\begin{equation*}
\mathbb{E}_{y\sim p_{\theta_0}(\cdot \mid x)} \sum_{n=1}^{L_y}\rho_{\theta}(y_n \mid y_{<n}, x)\log \frac{p_{\theta_0}(y_n \mid y_{<n}, x)}{q(y_n \mid y_{<n}, x)},
\end{equation*}
where $\rho_{\theta}=p_{\theta}(y_n \mid y_{<n}, x)/p_{\theta_0}(y_n \mid y_{<n}, x)$ is the importance sampling ratio. We provide a more detailed full derivation in Appendix \ref{App:OPD derivation}.

\section{OPD+: Correcting the Advantage Estimation for On Policy Distillaion}

\label{Sec: OPD+}

Note that in our derivation of OPD loss objective for KL divergence, it is assumed that the reward $\log \frac{p_{\theta}(y_n \mid y_{<n}, x)}{q(y_n \mid y_{<n}, x)}$ is fixed, and hence independent of $\theta$. Yet this leads to a potential mismatch between the true target and our loss objective: the gradient can be biased due to the stop gradient operations. 

\subsection{Beyond KL: $f$-divergence optimization for on-policy distillation}
We first present a generic framework of $f$-divergence optimization between the student and teacher policy, and derive the true gradient and the loss objective compared to those obtained from stop gradient operations.

\paragraph{$f$-divergence objective.} Let $f:(0,\infty)\to\mathbb{R}$ be a convex function satisfying $f(1)=0$, and define the token-level density ratio as $u_{\theta,n}:=q(y_n\,|\,y_{<n},x)/p_{\theta}(y_n\,|\, y_{<n},x).$
Consider the $f$-divergence objective sampling from the student model:
\begin{align}
J^f(\theta)
=
\mathbb{E}_{x\sim X}
\mathbb{E}_{y\sim p_{\theta}(\cdot\mid x)}
\sum_{n=1}^{L_y}
f\!\left(u_{\theta,n}\right).
\end{align}
Equivalently, minimizing $J^f(\theta)$ corresponds to minimizing the $f$-divergence between the student's and the teacher's next-token distribution along prefixes sampled from the student. 

\paragraph{Stop-gradient policy-gradient estimator.}
Define the reward as the negative divergence value:
\begin{align}
r_{\theta}^{f}(y_{\le n},x)
:=
-
f\!\left(u_{\theta,n}\right),
\end{align}
As in the earlier reverse-KL case, suppose we stop-gradient the reward term (denote $\theta_{-}$ for preventing gradient back propagation), the objective is:
$$
J_{\text{sg}}^f(\theta)
=
\mathbb{E}_{x\sim X,\ y\sim p_{\theta}(\cdot\mid x)}
\sum_{n=1}^{L_y}
\gamma^n r_{\theta_-}^{f}(y_{\le n},x).
$$
By REINFORCE \cite{williams1992simple}, $\nabla_{\theta}J_{\text{sg}}^f(\theta)
=$
$$
\mathbb{E}_{x\sim X,\ y\sim p_{\theta}(\cdot\mid x)}
\left[
\sum_{k=1}^{L_y}
\nabla_{\theta}\log p_{\theta}(y_k\mid y_{<k},x)
A^f_k
\right],
$$
where the reward-to-go advantage is $A_k^f:=\sum_{n=k}^{L_y}\gamma^{n-k} r_{\theta_-}^{f}(y_{\le n},x)$.
Taking the discount factor $\gamma = 0$ yields the immediate token-level reward
$A_k^f
=
r_{\theta_-}^{f}(y_{\le k},x)$.
Thus, the resulting gradient estimator of $J_{\text{sg}}^f$ is:
\begin{align}
\label{Stop gradient policy gradient}
\mathbb{E}_{y\sim p_{\theta}(\cdot\mid x)}
\sum_{n=1}^{L_y}
\nabla_{\theta}
\log p_{\theta}(y_n\mid y_{<n},x)
r_{\theta_-}^{f}(y_{\le n},x).
\end{align}




\subsection{OPD+: Corrected Formula for $f$-divergence Optimization.}
We derive our earlier formula by applying stop gradient to the reward term. However, the true reward 
$r_{\theta}^{f}(y_{\le n},x)
=
-
f\!\left(u_{\theta,n}\right)
$
does depend on $\theta$. 
So if we differentiate the true objective, we get two terms -- the {\em score-function term} and the {\em direct reward-gradient term}: $\nabla_{\theta}\mathbb{E}_{y_n\sim p_{\theta}(\cdot\mid y_{<n},x)}\left[r_{\theta}^{f}(y_{\le n},x)
\right]=$
$$
\begin{aligned}
&\mathbb{E}
[
r_{\theta}^{f}(y_{\le n},x)
\nabla_{\theta}
\log p_{\theta}(y_n| y_{<n},x)+
\nabla_{\theta} r_{\theta}^{f}(y_{\le n},x)
].
\end{aligned}
$$
The direct reward-gradient term is computed as:
\[
\begin{aligned}
\nabla_{\theta} r_{\theta}^{f}(y_{\le n},x)
=
u_{\theta, n}f'(u_{\theta, n})
\nabla_{\theta}
\log p_{\theta}(y_n\mid y_{<n},x).
\end{aligned}
\]
So the correct immediate token-level gradient is:
\[
\begin{aligned}
&\mathbb{E}_{y\sim p_{\theta}(\cdot\mid x)}
\sum_{n=1}^{L_y}
\nabla_{\theta}\log p_{\theta}(y_n\mid y_{<n},x)w_f(u_{\theta, n}),
\end{aligned}
\]
where
$
w_f(u_{\theta, n})
=
-f(u_{\theta, n})
+
u_{\theta, n}f'(u_{\theta, n}).
$
Thus, for a general $f$-divergence, the correct advantage function 
is not simply $r_{\theta}^{f}(y_{\le n},x)=-f(u_{\theta, n})$: there is an additional term of $u_{\theta, n}f'(u_{\theta, n})$ 
due to the reward dependency on policy parameter $\theta$, and we name our method as OPD+. To implement OPD+, only one line change of code is needed.

\paragraph{What is special about Reverse KL?}
So is the exisiting formula of reverse KL misused when we do not apply stop grad? Theoretically, the answer is {\bf no}. 
Reverse KL is a special instance of $f$-divergence with
$f(u)=-\log u$.
As a result,
\[
w_f(u)
=
f(u)-uf'(u)
=
-\log u + 1.
\]
The additional $+1$ term is a constant baseline, and hence vanishes under the score identity
$
\mathbb{E}_{y_n\sim p_{\theta}}
\left[
\nabla_{\theta}
\log p_{\theta}(y_n\mid y_{<n},x)
\right]
=
0$.
Thus, the direct reward-gradient correction for reverse KL reduces to a constant baseline, and is {\em hidden} in the stop-gradient policy optimization.
For a general $f$, the term $uf'(u)$ is not a constant, 
so the stop gradient operation leads to a bias. 
See Table \ref{tab:advantage for f divergence} for comparisons of different divergences.

\begin{table}[!th]
\centering
\renewcommand{\arraystretch}{0.85}
\resizebox{\columnwidth}{!}{%
\begin{tabular}{lcc}
\toprule
\textbf{Divergence} 
& \textbf{Stop Grad Adv.} 
& \textbf{Corrected Adv.} \\
\midrule
Forward KL 
& $-u \ln u$ 
& $u$ \\

Reverse KL 
& $\ln u$ 
& $\ln u - 1$ \\

JSD 
& $-\frac{1}{2}\left[u \ln u - (1+u)\ln\frac{1+u}{2}\right]$ 
& $\frac{1}{2}\ln\frac{1+u}{2}$ \\
\bottomrule
\end{tabular}
}
\caption{Advantage Design for On-Policy Distillation.}
\label{tab:advantage for f divergence}
\end{table}

In addition, despite the theoretical coincidence, we observe the empirical performance gain in adding the baseline term even for KL divergence. 

\begin{table*}[!th]
\centering
\resizebox{0.9\textwidth}{!}{
\begin{tabular}{llcccccc}
\toprule
& & \multicolumn{4}{c}{Competition Math} & Agentic & Self-Distillation \\
\cmidrule(lr){3-6}\cmidrule(lr){7-7}\cmidrule(lr){8-8}
Div. Type & Advantage & AIME 24 & AIME 25 & HMMT 25 & Avg. & Tool Use & AIME 24 \\
\midrule
\multicolumn{2}{l}{SFT-init student} 
& 50.94 & 38.61 & 25.52 & 38.36 & 24.63 & 48.89 \\
\midrule
\multirow{2}{*}{Forward KL} 
& \opd{}: $-f(u)$ 
& \basegray{50.94} & \basegray{38.61} & \basegray{25.52} & \basegray{38.36} & \basegray{24.63} & \basegray{48.89} \\
& \methodname{}: $w(u)$ 
& 59.79 & 46.88 & 32.50 & 46.39 & 59.01 & \underline{50.83} \\
\midrule
\multirow{2}{*}{JSD} 
& \opd{}: $-f(u)$ 
& \basegray{50.94} & \basegray{38.61} & \basegray{25.52} & \basegray{38.36} & \basegray{24.63} & 49.17 \\
& \methodname{}: $w(u)$ 
& 63.12 & \textbf{50.42} & 33.02 & 48.85 & \underline{60.29} & 49.17 \\
\midrule
\multirow{2}{*}{Reverse KL} 
& \opd{}: $-f(u)$ 
& \underline{64.27} & 48.54 & \textbf{34.06} & \underline{48.96} & 59.97 & 50.00 \\
& \methodname{}: $w(u)$ 
& \textbf{64.69} & \underline{48.85} & \underline{33.44} & \textbf{48.99} & \textbf{61.44} & \textbf{53.33} \\
\bottomrule
\end{tabular}
}
\caption{Best-checkpoint avg@32 accuracy (\%, $\uparrow$) across math, tool-use, and self-distillation tasks. We compare standard OPD using $-f(u)$ with the corrected OPD+ using $w(u)$ across different $f$-divergences. Bold/underline mark the best/second-best results within each task. Grey cells indicate runs whose performance fell below the SFT-init student baseline; their displayed value is clipped to the baseline performance. OPD+ eliminates the training collapse seen in standard Forward KL and JSD. Self-Distillation corresponds to results of \textsc{Qwen3-1.7B}.}
\label{tab:main}
\end{table*}

\section{Experiments}

\label{Sec: Exp}

We conduct experiments to showcase the advantage of our \methodname{}. We test our method across two distinct training paradigms spanning \textbf{\textit{competition mathematics}} and \textbf{\textit{agentic tool use}}, and use either a separate, stronger model (OPD) or the model itself with privileged ground-truth solutions as the teacher (OPSD, \cite{zhao2026self}).


\paragraph{Models, Tasks and Metrics.}
We implement \textsc{OPD} and OPD+ across multiple scales of the \textsc{Qwen3} model family (\textsc{Qwen3-1.7B} and \textsc{8B}). 
We benchmark our approach on two demanding application domains, evaluating performance via tasks that assess distinct capability dimensions.  For \textbf{\textit{competition mathematics}}, we distill on DeepMath \cite{deepmath} rollouts and evaluate on high-bar reasoning benchmarks, specifically AIME 24, AIME 25, and HMMT 25. For \textbf{\textit{tool use}}, we distill on the ToolUse task \cite{tang2023toolalpaca} and evaluate on its 68-problem test split, where a teacher reference score is also available for baseline comparison. For both tasks, the evaluation metric is \texttt{avg@32}, which is the mean per-sample score as an estimate of single-sample accuracy. More details could be found in Appendix \ref{sec:experimental details}.

\begin{figure*}[!th]
  \centering
    \includegraphics[width=0.9\linewidth]{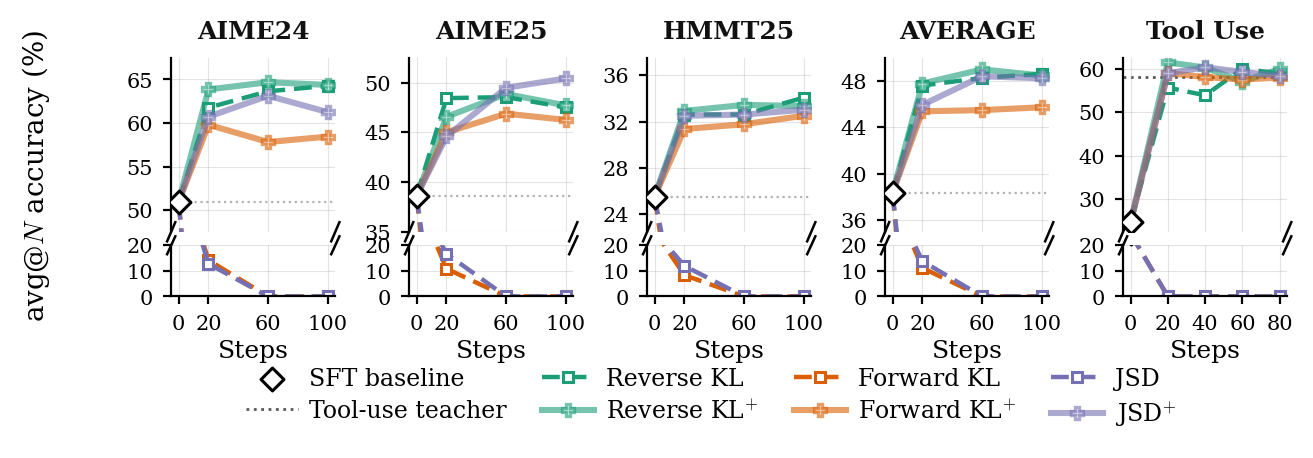}
    \caption{Training dynamics of OPD vs. OPD+ across different steps: we report average accuracy \texttt{avg@32} ($\uparrow$) on three math benchmarks (AIME 24, AIME 25, HMMT 25), their average, and an agentic Tool Use task (rightmost panel). Solid lines with markers ($+$) represent our corrected OPD+ variants ($w(u)$), while dashed lines ($\square$) denote standard OPD baselines ($-f(u)$). OPD+ consistently accelerates early convergence or prevents catastrophic training collapse.}
    \label{fig:opd_math}
\end{figure*}

\subsection*{Main Results}
We report in Table \ref{tab:main}, the best-checkpoint performance achieved by each method across the evaluated tasks. Our results present a striking challenge to the prevailing post-training paradigm, which universally favors reverse KL due to its empirical stability while avoiding forward KL and JSD.

\paragraph{The Salvation of Forward KL and JSD.} Under the standard bare-cost advantage $-f(u)$, both Forward KL and JSD exhibit catastrophic collapse, failing to match even the SFT-initialized baseline on competition mathematics, and dropping to  $0.00\%$ accuracy on tool use. 
In contrast, our analytic correction, OPD+, completely eliminates this divergence-dependent failure mode. By replacing $-f(u)$ with the gradient-faithful weight $w(u)$, OPD+ rescues Forward KL and JSD from collapse, elevating their performance to be fully competitive with or superior to the Reverse KL benchmark.

\paragraph{Surpassing the Established Paradigm.} Crucially, the empirical gains of OPD+ extend beyond merely stabilizing previously un-trainable objectives; it fundamentally expands the performance ceiling of on-policy distillation:

$\diamondsuit$ \textbf{The Superiority of $\text{JSD}^+$}: While JSD was previously considered completely non-viable on-policy, $\text{JSD}^+$ not only matches the conventional Reverse KL paradigm on mathematical averages but substantially outperforms it on the challenging AIME 25 benchmark ($50.42\%$ vs. $48.54\%$). More importantly, on the Tool Use task, $\text{JSD}^+$ establishes a new performance peak of $60.29\%$, surpassing the standard Reverse KL baseline ($59.97\%$).
    
$\diamondsuit$ \textbf{Bootstrapping Reverse KL Itself}: Even within the native domain of Reverse KL, injecting our derived formulation yields strict empirical improvements. Compared to the standard Reverse KL baseline, $\text{Reverse KL}^+$ achieves consistent performance boosts across three out of the four evaluated benchmarks: AIME 24 ($64.69\%$ vs. $64.27\%$), AIME 25 ($48.85\%$ vs. $48.54\%$), and Tool Use ($61.44\%$ vs. $59.97\%$). These findings demonstrate that the historical underperformance of alternative $f$-divergences is entirely a side-effect of a mis-specified gradient coefficient, rather than an intrinsic limitation of the divergences themselves. When optimized faithfully, they offer powerful alternatives that outcompete the current industry standard.

\paragraph{OPSD+ also outperforms OPSD baselines.} We observe that the our method still demonstrate better performance compared to the baseline approaches in the OPSD setup when the teacher model is model itself with privileged information. As in the last row of Table \ref{tab:main}, when using \textsc{Qwen3 1.7B instruct} as the base model, the improved versions all outperforms the vanilla stop gradient version, and the improved reverse KL showcases the best performance among different divergences.

\section{Conclusions and Discussion}

\label{Sec: Conclusion}

In this paper, we investigate on the two key parts among the design space of OPD: the divergence and the choice of reward/advantage. We provide OPD+, the generic framework upon $f$-divergence optimization and the corrected formulas compared to vanilla policy gradient based on the stop gradient operations. Our techniques provide both more mathematically correct and empirically well-performed algorithms, challenging the existing practice of divergence choice and design space of the reward. We believe that our paper could trigger further research on better design of advantage function of OPD type methods.

\section*{Acknowlegdment}
Tang is supported by NSF CAREER Award DMS-2538791 and the Tang
Family Assistant Professorship.
Tang, Yao and Zhao are part of a Columbia-CityU/HK collaborative project that is supported by InnoHK Initiative, The Government of the HKSAR and the AIFT Lab. 

\section*{Limitations}
In this paper, we did not do a detailed model scaling analysis to evaluate the sensitivity of our algorithms and baseline methods to the model scale. We also limit the analysis of our experiments to the Qwen3 model families, so the results beyond Qwen model families will be helpful to see the generalization capability of our findings. We also do not compare the performance of single token performance with logit distillation performance for various divergences, which we would pursue as future work to better understand the tradeoffs.

\section*{Ethical Considerations}
Our work primarily targets at advancing the field of Machine Learning and NLP. There are no ethical considerations that we believe are worth mentioning.

\bibliography{custom}

\clearpage

\appendix

\section{Related Works}
\label{App: related works}

\textbf{$f$-divergence in post training generative models}. Statistical divergence beyond KL has been studied in the context of RLHF as regularizers for post-training LLMs \cite{wang2024beyond,laidlaw2025correlated, zhang2025design}. 
Relevant extensions has also been studied for fine-tuning diffusion/flow models \cite{tang2024fine, de2026flow, guo2026improved}. The possible biasedness of directly differentiated KL gradient estimator has been discussed in \cite{tang2025few, shah2025comedy}. Comparatively in OPD, divergence choice acts as the primary reward, which is of more crucial importance for downstream tasks than for alignment/reasoning regularizors.\\

\noindent \textbf{OPD}. \citep{agarwal2024policy} proposed GKD for LLMs, which corresponds to a supervised learning form of OPD and also proposes using various $f$-divergence in the same off-policy setup. \cite{lu2025onpolicydistillation} discussed the RL like variant of OPD for KL divergence, including the application in LLM reasoning and personalizations. See \cite{song2026survey} for a comprehensive survey on the followup works and recent progress. More recently, OPSD \cite{zhao2026self} type of methods generalize the setups of OPD to using the models itself with priviledge information as the teacher model. Concurrent work like SDPO \cite{hubotter2026reinforcement} and SDFT \cite{shenfeld2026self} proposed similar ideas by letting the model condition on environment feedback. 
In the same spirit, \cite{song2026expanding} also discussed using the modified solutions from LLMs conditioned on text feedback to provide similar dense supervision. 

\section{OPD Derivations}
\label{App:OPD derivation}
We provide a more detailed derivation of OPD, and comparison of RL like loss in our paper to a supervised learning which is used in practice among different code repositories.

\paragraph{RL Loss.} To simplify the notation, we now define the per-token reward as the negative sampled reverse-KL term:
\[
r_{\theta}(y_{\le n},x)
=
\log q(y_n\mid y_{<n},x)
-
\log p_{\theta}(y_n\mid y_{<n},x).
\]
Therefore, minimizing the reverse-KL objective
\[
J_{\mathrm{KL-rev}}(\theta)
=
\mathbb{E}_{x\sim X, y\sim p_{\theta}(\cdot\mid x)}
\sum_{n=1}^{L_y}
\log
\frac{
p_{\theta}(y_n\mid y_{<n},x)
}{
q(y_n\mid y_{<n},x)
}
\]
is equivalent to maximizing
\[
J_{\mathrm{reward}}(\theta)
=
\mathbb{E}_{x\sim X}
\mathbb{E}_{y\sim p_{\theta}(\cdot\mid x)}
\sum_{n=1}^{L_y}
r_{\theta}(y_{\le n},x).
\]

As a common practice, the reward term is usually stop-gradient'ed in the policy-gradient estimator. We write
\[
r_{\theta_-}(y_{\le n},x)
:=
\log q(y_n\mid y_{<n},x)
-
\log p_{\theta_-}(y_n\mid y_{<n},x),
\]
where $\theta_-$ denotes the stop-gradient version of $\theta$. The corresponding RL-style objective is
\[
J_{\text{sg}}(\theta)
=
\mathbb{E}_{x\sim X}
\mathbb{E}_{y\sim p_{\theta}(\cdot\mid x)}
\sum_{n=1}^{L_y}
r_{\theta_-}(y_{\le n},x).
\]
By the REINFORCE formula, its gradient is
\[
\nabla_{\theta}J_{\text{sg}}(\theta)
=
\mathbb{E}_{y\sim p_{\theta}(\cdot\mid x)}
\left[
\sum_{k=1}^{L_y}
\nabla_{\theta}
\log p_{\theta}(y_k\mid y_{<k},x)
A_k
\right].
\]
where the advantage function is defined as
$
A_k
=
\sum_{n=k}^{L_y}
r_{\theta_-}(y_{\le n},x).
$
However, in token-level on-policy distillation, one commonly uses discount factor $\gamma=0$, which is reported by \cite{lu2025onpolicydistillation} that there is no improved performance from $\gamma>0$, despite that they are more mathematically correct. This thus replaces the full future reward-to-go by the immediate reward:
\[
A_k
=
\sum_{n=k}^{L_y}\gamma^{n-k}r_{\theta_-}(y_{\le n},x)
\Longrightarrow
A_k
=
r_{\theta_-}(y_{\le k},x),
\]
when $\gamma=0$. Therefore the $\gamma=0$ policy-gradient estimator becomes
\[
\begin{aligned}
&\nabla_{\theta}J_{\text{sg}}^{\gamma=0}(\theta)
=\\
&\mathbb{E}_{x\sim X}
\mathbb{E}_{y\sim p_{\theta}(\cdot\mid x)}
\sum_{n=1}^{L_y}
\nabla_{\theta}
\log p_{\theta}(y_n\mid y_{<n},x)
r_{\theta_-}(y_{\le n},x).
\end{aligned}
\]
Equivalently, the corresponding minimization loss is
\[
\mathcal L_{\mathrm{RL}}(\theta)
=
-
\mathbb{E}_{x\sim X}
\mathbb{E}_{y\sim p_{\theta}(\cdot\mid x)}
\sum_{n=1}^{L_y}
r_{\theta_-}(y_{\le n},x),
\]
In practice, trajectories are usually sampled from a frozen behavior model $p_{\theta_0}$. For each sampled token $y_n\sim p_{\theta_0}(\cdot\mid y_{<n},x)$, define
\[
r_{\theta_0}(y_{\le n},x)
=
\log q(y_n\mid y_{<n},x)
-
\log p_{\theta_0}(y_n\mid y_{<n},x),
\]
and the token-level importance ratio
\[
\rho_n(\theta)
=
\frac{
p_{\theta}(y_n\mid y_{<n},x)
}{
p_{\theta_0}(y_n\mid y_{<n},x)
}.
\]
The practical importance-sampling RL loss is then
\[
\begin{aligned}
&\mathcal L_{\mathrm{RL}}(\theta;\theta_0)
=\\
&-
\mathbb{E}_{x\sim X}
\mathbb{E}_{y\sim p_{\theta_0}(\cdot\mid x)}
\sum_{n=1}^{L_y}
\rho_n(\theta)
r_{\theta_0}(y_{\le n},x),
\end{aligned}
\]
with its gradient being:
\[
\begin{aligned}
&
\nabla_{\theta}\mathcal L_{\mathrm{RL}}(\theta;\theta_0)
=-
\mathbb{E}_{y\sim p_{\theta_0}(\cdot\mid x)}
\sum_{n=1}^{L_y}
\rho_n(\theta)
r_{\theta_0}(y_{\le n},x)\\
&\nabla_{\theta}
\log p_{\theta}(y_n\mid y_{<n},x).
\end{aligned}
\]
At $\theta=\theta_0$, we have $\rho_n(\theta_0)=1$, and hence
\[
\begin{aligned}
&\nabla_{\theta}\mathcal L_{\mathrm{RL}}(\theta;\theta_0)
\bigg|_{\theta=\theta_0}
=
-
\mathbb{E}_{x\sim X}
\mathbb{E}_{y\sim p_{\theta_0}(\cdot\mid x)}
\sum_{n=1}^{L_y}\\
&\nabla_{\theta}
\log p_{\theta}(y_n\mid y_{<n},x)
\left[
\log \frac{q(y_n\mid y_{<n},x)}{p_{\theta_0}(y_n\mid y_{<n},x)}
\right].
\end{aligned}
\]
\paragraph{Supervised Learning Loss.} Next we consider the supervised-learning style loss on the same frozen on-policy prefix data. After sampling $y\sim p_{\theta_0}(\cdot\mid x)$, we freeze the prefixes $(y_{<n},x)$ and directly minimize the full-vocabulary reverse KL:
\[
\begin{aligned}
&\mathcal L_{\mathrm{supKL}}(\theta;\theta_0)
=
\mathbb{E}_{x\sim X}
\mathbb{E}_{y\sim p_{\theta_0}(\cdot\mid x)}
\sum_{n=1}^{L_y}\\
&D_{\mathrm{KL}}
\left(
p_{\theta}(\cdot\mid y_{<n},x)
\|
q(\cdot\mid y_{<n},x)
\right).
\end{aligned}
\]
Expanding over the vocabulary $\mathcal V$,
\[
\begin{aligned}
&\mathcal L_{\mathrm{supKL}}(\theta;\theta_0)
=
\mathbb{E}_{x\sim X}
\mathbb{E}_{y\sim p_{\theta_0}(\cdot\mid x)}
\sum_{n=1}^{L_y}\\
&\sum_{a\in\mathcal V}
p_{\theta}(a\mid y_{<n},x)
\log
\frac{
p_{\theta}(a\mid y_{<n},x)
}{
q(a\mid y_{<n},x)
}.
\end{aligned}
\]
Its gradient is $\nabla_{\theta}\mathcal L_{\mathrm{supKL}}(\theta;\theta_0)
=$
\[
\begin{aligned}
&\mathbb{E}_{x\sim X}
\mathbb{E}_{y\sim p_{\theta_0}(\cdot\mid x)}
\sum_{n=1}^{L_y}\mathbb{E}_{a\sim p_{\theta}(\cdot\mid y_{<n},x)}\\
&\left[
\nabla_{\theta}
\log p_{\theta}(a\mid y_{<n},x)
\log
\frac{
p_{\theta}(a\mid y_{<n},x)
}{
q(a\mid y_{<n},x)
}
\right].
\end{aligned}
\]
The extra $+1$ term from differentiating $p_\theta\log p_\theta$ theoretically vanishes because of score identity. Evaluating the supervised-KL gradient at $\theta=\theta_0$ gives $\nabla_{\theta}\mathcal L_{\mathrm{supKL}}(\theta;\theta_0)
\bigg|_{\theta=\theta_0}=$
\[
\begin{aligned}
&
\mathbb{E}_{x\sim X}
\mathbb{E}_{y\sim p_{\theta_0}(\cdot\mid x)}
\sum_{n=1}^{L_y}
\mathbb{E}_{a\sim p_{\theta_0}(\cdot\mid y_{<n},x)}\\
&\left[
\nabla_{\theta}
\log p_{\theta}(a\mid y_{<n},x)
\left(
\log \frac{p_{\theta_0}(a\mid y_{<n},x)}{q(a\mid y_{<n},x)}
\right)
\right]_{\theta=\theta_0}.
\end{aligned}
\]
The RL surrogate above is exactly a Monte Carlo estimator of this gradient, using the sampled token $a=y_n$. Therefore,
\[
\nabla_{\theta}\mathcal L_{\mathrm{RL}}(\theta;\theta_0)
\bigg|_{\theta=\theta_0}
=
\nabla_{\theta}\mathcal L_{\mathrm{supKL}}(\theta;\theta_0)
\bigg|_{\theta=\theta_0}
\]
in expectation over the sampled token $y_n\sim p_{\theta_0}(\cdot\mid y_{<n},x)$.\\

However, the two losses are not globally identical. Away from $\theta_0$, the supervised KL gradient uses the current coefficient
\[
\log p_{\theta}(a\mid y_{<n},x)-\log q(a\mid y_{<n},x),
\]
whereas the RL surrogate uses the frozen rollout coefficient
\[
\log p_{\theta_0}(a\mid y_{<n},x)-\log q(a\mid y_{<n},x).
\]
Thus, in general,
\[
\nabla_{\theta}\mathcal L_{\mathrm{RL}}(\theta;\theta_0)
\neq
\nabla_{\theta}\mathcal L_{\mathrm{supKL}}(\theta;\theta_0)
\qquad
\text{for } \theta\neq \theta_0.
\]
The equivalence is only a local gradient equivalence at the rollout policy $\theta_0$.

\section{More Experimental Details and Analysis}
\label{sec:experimental details}
\subsection{More details on Experimental Setup}
In \textbf{\textit{OPD}}, we distill a Qwen3-8B teacher into a Qwen3-8B-Base student model which has been initialized by supervised fine-tuning (SFT). For competition mathematics, the student trains for 100 on-policy steps (with a total batch size of 64 groups). For tool use, distillation proceeds for 80 steps (256 rollouts per step) under identical architecture settings. Notably for math reasoning tasks, we used the default hyperparameter setup of the KL divergence baseline already tuned and recommended by tinker of the Thinking Machine Labs, and our OPD+ achieves better results with simple one line change of code under the same hyperparameter setups. It is possible that better results can be achieved through more detailed sweeping of hyperparameters.\\

\noindent In \textbf{\textit{OPSD}}, we a use the same model to act as both student and teacher simultaneously by varying the conditioning context: the teacher policy operates under an RL setting by conditioning on privileged information (the problem and its reference solution), while the student policy observes only the problem statement. We use \textsc{Qwen3}-1.7B with a fixed LoRA teacher (rank 64), a learning rate of $5\times10^{-6}$, and a 1024-token generation budget. 

\subsection{Other OPD Experimental Results Analysis}
To understand the mechanics underlying the stability and performance improvement of OPD+, we trace the single-sample accuracy (\texttt{avg@32}) and generation properties across the full training trajectory.

\paragraph{Convergence Acceleration via Variance Control.} 
While our analysis in Section 3.2 shows that the bare cost and analytic weight for Reverse KL induce identical gradients in expectation ($w = g + 1$), the training dynamics reveal a highly consequential empirical phenomenon: $\text{Reverse KL}^+$ (solid green line) exhibits significantly greater \textbf{sample efficiency and early convergence velocity} than standard Reverse KL (dashed green line). 

As shown in Figure \ref{fig:opd_math}, on AIME 24, $\text{Reverse KL}^+$ undergoes an immediate performance leap, approaching its peak accuracy ($\sim 64\%$) by step 20, whereas the uncorrected baseline lags noticeably behind at the same interval. We observe a matching trend on HMMT 25, where the solid green curve reaches its peak plateau at step 20 while the standard paradigm slowly climbs to catch up in later stages. This acceleration is even more pronounced in the Tool Use task (last figure of Figure \ref{fig:opd_math}), where the solid $\text{Reverse KL}^+$ curve maintains a dominant upper bound over its dashed counterpart throughout the crucial step 20--40 training window. 

In the policy gradient framework, shifting a per-token score function by a state-independent constant $+1$ acts as an exact, analytically derived variance-reduction baseline. Our training curves substantiate that this built-in baseline mechanism effectively stabilizes early gradient updates, converting directly into faster optimization rates and superior sample efficiency during post-training.

\paragraph{Sustained Optimization Resilience of $\text{JSD}^+$.}
Beyond accelerating convergence, the corrected JSD ($\text{JSD}^+$, solid purple line) exhibits exceptional optimization stability and late-stage stamina. While the stop gradient JSD baseline (dashed purple line) suffers from immediate and irreversible collapse to $0\%$ accuracy across all tasks, $\text{JSD}^+$ not only completely averts this failure mode but continuously expands its performance ceiling as training progresses. 

This behavior is most prominently illustrated on AIME 2025 (Figure \ref{fig:opd_math}): while other competitive variants (including Reverse KL) plateau or begin to slightly degrade after step 60, the trajectory of $\text{JSD}^+$ maintains an upward trajectory, achieving the absolute highest accuracy ($50.42\%$) at step 100. This sustained growth highlights that symmetrically balancing the properties of both forward and reverse distributions, when mathematically formalized via the proper analytic weight $w(u)$, yields a inherently robust training signal that actively counteracts late-stage overfitting or policy degradation.

\subsection{Other OPSD Experimental Figures}
We also report the concrete training curves of OPSD experiments, on AIME24.
\begin{figure}[!th]
  \centering
    \includegraphics[width=0.9\linewidth]{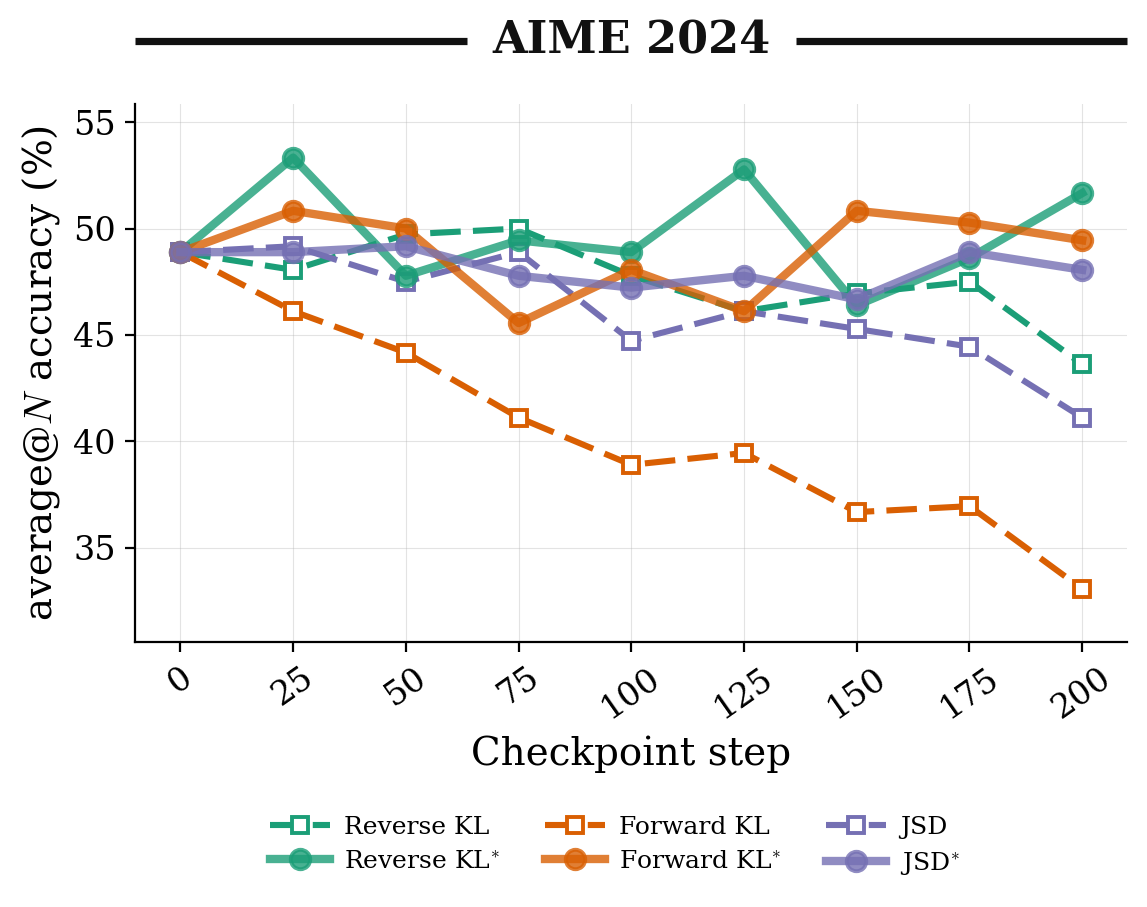}
    \caption{Training dynamics of OPSD vs. OPSD+ across different steps: OPSD+ achieves both better highest performance and clearly prevents catastrophic training collapse.}
    \label{fig:opsd_aime24}
\end{figure}
\end{document}